\documentclass[conference]{IEEEtran}
\IEEEoverridecommandlockouts

\usepackage{cite}
\usepackage{amsmath,amssymb,amsfonts}
\usepackage{algorithmic}
\usepackage{graphicx}
\usepackage{textcomp}
\usepackage{xcolor}
\usepackage{url}

\usepackage{adjustbox} 
\usepackage{array}

\def\BibTeX{{\rm B\kern-.05em{\sc i\kern-.025em b}\kern-.08em
    T\kern-.1667em\lower.7ex\hbox{E}\kern-.125emX}}

\begin{document}

\title{Machine Learning–Based Pre-Test Risk Stratification for PCR-Confirmed Chlamydia Using Patient-Reported Data and Urine Biomarkers\\

}

\author{
\IEEEauthorblockN{
Mehrab Mahdian$^{1,*}$,
Marko Lehes$^{2,3}$,
Katrin Krolov$^{4}$,
Tamas Pardy$^{1,2}$
}
\IEEEauthorblockA{$^{1}$\textit{Tallinn University of Technology (TalTech)}, Tallinn, Estonia}
\IEEEauthorblockA{$^{2}$\textit{Selfdiagnostics Deutschland GmbH}, Tallinn, Estonia}
\IEEEauthorblockA{$^{3}$\textit{Capitol Technology University}, Laurel, MD, USA}
\IEEEauthorblockA{$^{4}$\textit{University of Tartu}, Tartu, Estonia}
\IEEEauthorblockA{mehrab.mahdian@taltech.ee, marko.lehes@selfdiagnostics.com, katrin.krolov@ut.ee, tamas.pardy@taltech.ee}
\IEEEauthorblockA{$^{*}$Corresponding author: mehrab.mahdian@taltech.ee}
}
\maketitle

\begin{abstract}
Early identification of individuals at elevated risk of Chlamydia trachomatis infection may enable optimal use of molecular testing in resource-aware screening. We evaluate the feasibility of pre-test risk stratification (PTRS) using machine-learning models trained on routinely available, non-invasive clinical data. 

A curated dataset of 93 urine samples with PCR reference labels was analyzed using three feature groups: patient-reported history and symptoms, urine biomarkers from standard urinalysis, and their combination. Five supervised classifiers were evaluated using stratified 5-fold cross-validation with out-of-fold probability estimates. Performance was assessed using area under the receiver operating characteristic curve (AUC) and threshold-dependent metrics, with uncertainty quantified via bootstrap confidence intervals. 

Models using only patient-reported data showed moderate discrimination (AUC up to 0.72). Urine biomarker-based models demonstrated slightly lower peak discrimination but more consistent performance, with ensemble methods yielding the strongest results. Combining feature groups marginally increased the peak AUC and reduced performance variability across models, indicating improved robustness.

Findings indicate that urine biomarkers provide a reliable predictive signal for PTRS that is complementary to patient-reported information, while feature integration enhances robustness. This work supports the integration of non-invasive, routinely available information for PTRS into screening workflows, including decentralized or home-based PCR contexts, to optimize testing prioritization.
\end{abstract}

\begin{IEEEkeywords}
Chlamydia trachomatis, machine learning, risk stratification, urine biomarkers, clinical decision support, pre-test screening.

\end{IEEEkeywords}

\section{Introduction}

\textit{Chlamydia trachomatis} is one of the most prevalent bacterial sexually transmitted infections worldwide, with an estimated 128.5 million new infections annually among adults aged 15–49 years, and it represents a significant public health burden in part because many infected individuals are asymptomatic \cite{WHOChlamydiaFactSheet, Chang2023ChlamydiaEditorial}. A substantial proportion of infections remain asymptomatic or present with nonspecific clinical signs, contributing to delayed diagnosis and ongoing transmission \cite{Peeling2017AsymptomaticSTI}. Nucleic acid amplification tests (NAATs), including PCR-based assays, are regarded as the diagnostic gold standard due to their high analytical sensitivity and specificity \cite{CDC2021ChlamydiaGuidelines, Gaydos2017NAATReview}. However, universal PCR testing is costly, resource-intensive, and may be impractical in settings with limited laboratory capacity or high screening demand \cite{Chakraborty2024DemocratizingNAAT}.

As a result, there is ongoing interest in pre-test risk stratification strategies that can help prioritize individuals for confirmatory testing \cite{WHO2025STIGuidance, Bilal2021STIPredictionML}. Patient-reported information, such as sexual history, prior sexually transmitted infections, and self-reported symptoms, represents a low-cost, non-invasive source of data available before laboratory testing \cite{Hao2025STIRiskPrioritization}. Epidemiological studies have demonstrated associations between these factors and chlamydia risk, yet their standalone predictive value at the individual level remains limited \cite{vanKlaveren2016ChlamydiaPrediction, Jensen2025ChlamydiaRiskFactors}.  In parallel, routinely collected urine biomarkers provide objective biological information that may reflect underlying infection or inflammation, but their added value relative to patient-reported data is not well characterized \cite{Sequeira-Antunes2023UrinaryBiomarkers}.

Machine learning (ML) methods have increasingly been explored for clinical risk prediction and decision support, including applications in sexually transmitted disease screening \cite{Rajkomar2019MLHealthcare, Bilal2021MLSTDIPredict}. While these approaches offer flexibility in modeling nonlinear relationships, their application to small clinical datasets raises concerns regarding overfitting, data leakage, and limited interpretability. Many prior studies emphasize predictive performance without sufficiently isolating the contribution of different feature groups or quantifying uncertainty, limiting their utility for cautious clinical translation.

In this study, we investigate the feasibility of predicting PCR-confirmed \textit{Chlamydia trachomatis} infection, using ML models trained on (i) patient-reported history and symptoms, (ii) urine biomarkers, and (iii) their combination. Rather than proposing a diagnostic replacement for PCR, the objective is to systematically compare these feature groups under a controlled evaluation framework and assess their relative predictive utility for pre-test triage. By employing predefined feature sets, conservative model complexity, out-of-fold prediction, and bootstrap-based uncertainty estimation, this work aims to provide a transparent and reproducible baseline for exploratory risk stratification in resource-aware screening contexts.

\section{Materials and Methods}

\subsection{Dataset and Study Population}
The dataset was assembled from multiple clinical studies and assay evaluation cohorts conducted by Selfdiagnostics Deutschland GmbH between 2018 and 2019, comprising urine samples collected in routine sexual health screening contexts. Samples originated from three primary sources: a university hospital–based clinical study conducted in Tartu, Estonia (UT clinical study 2018); a CE-marking–related clinical evaluation conducted in Leipzig, Germany (Leipzig CE study 2019); and a smaller set of urine samples collected during beta-stage LAMP assay evaluations performed in Tartu, Estonia.

Urine samples were collected as part of standard-of-care diagnostic workflows. Visual assessment of samples was performed at the time of collection and recorded as free-text descriptors. Reference infection status for \textit{Chlamydia trachomatis} was determined using PCR-based testing with the Roche cobas assay.

\subsection{Data Curation and Feature Engineering}

Raw clinical data were curated through a structured, rule-based preprocessing pipeline designed to maximize interpretability while minimizing information leakage. Initial data loading was performed from a formatted spreadsheet, after which records failing internal quality control were excluded based on control flags provided in the dataset. Only samples with valid control status were retained for subsequent analysis. Samples originating from beta-stage LAMP assay evaluations were excluded during data cleaning due to incomplete and non-uniform metadata.

Feature engineering was performed using predefined rules applied \textit{a priori}. Variables exhibiting no variance, excessive missingness, or latent diagnostic content were excluded prior to modeling. Where applicable, redundant fine-grained indicators were aggregated into higher-level proxy variables to reduce sparsity and improve robustness. Only non-invasive features available prior to laboratory testing were retained.

Qualitative visual assessments of urine samples were originally recorded as free-text descriptions. These entries were normalized using deterministic text parsing rules to extract two categorical features: sample color (\textit{light}, \textit{average}, \textit{dark}, \textit{unknown}) and sample cloudiness (\textit{cloudless}, \textit{cloudy}, \textit{very cloudy}, \textit{unknown}). Parenthetical comments and inconsistent separators were removed prior to normalization. The original free-text field was discarded after encoding.

In the following, engineered variables were organized \textit{a priori} into three semantically distinct feature groups to enable controlled comparison of different information sources: (F1) patient-reported history and symptoms, (F2) urine-derived biomarkers from routine urinalysis, and (F3) a combined feature set integrating F1 and F2. This grouping strategy was used consistently throughout model training and evaluation.

\subsubsection{Patient-Reported History and Symptoms (F1)}

Patient-reported features included demographic information, sexual history, prior sexually transmitted disease (STD) history, medication use, chronic disease indicators, and reported symptoms. Binary encoding was applied to categorical variables, and gender was mapped to a binary indicator. 

The final F1 feature set included gender, age, new sexual partner status, recent unprotected intercourse, unprotected intercourse with a new partner within the past 12 months, prior STD diagnosis, recent painkiller use, presence of chronic disease, and self-reported urogenital symptoms, including dysuria, abnormal genital discharge, intermenstrual bleeding, genital irritation or itching, and urinary urgency.

\subsubsection{Urine Biomarkers (F2)}

Urine biomarker features consisted of routinely measured quantitative parameters derived from standard urine analysis. Semi-quantitative measurements expressed using inequality symbols were converted to numeric values using predefined parsing rules, and biomarkers with no variance or excessive missingness across samples were excluded. This feature group represents objective biological signal derived from urine analysis.

The final F2 feature set comprised leukocyte count, bilirubin concentration, protein concentration, urine specific gravity, pH, ascorbic acid, microalbumin, calcium, and creatinine.

\subsubsection{Combined Feature Set (F3)}

The combined feature set (F3) integrated all patient-reported features from F1 with urine biomarkers from F2, enabling evaluation of complementary information across feature groups.

After feature-level exclusions were applied within each feature group, sample-level missingness was addressed to ensure consistent model input across feature sets. Rows containing missing values in the final combined feature set were excluded, and the same row indices were applied to F1 and F2, resulting in aligned datasets with identical sample counts across all feature groups. Categorical variables were encoded using binary or one-hot encoding as appropriate, with reference categories dropped to avoid collinearity. No imputation was performed.

This feature engineering strategy prioritizes transparency, reproducibility, and clinical plausibility, enabling controlled comparison of patient-reported and biomarker-based information under identical evaluation conditions.

\subsection{Study Cohort After Data Cleaning}

\begin{figure}[htbp]
\centering
\includegraphics[width=\columnwidth]{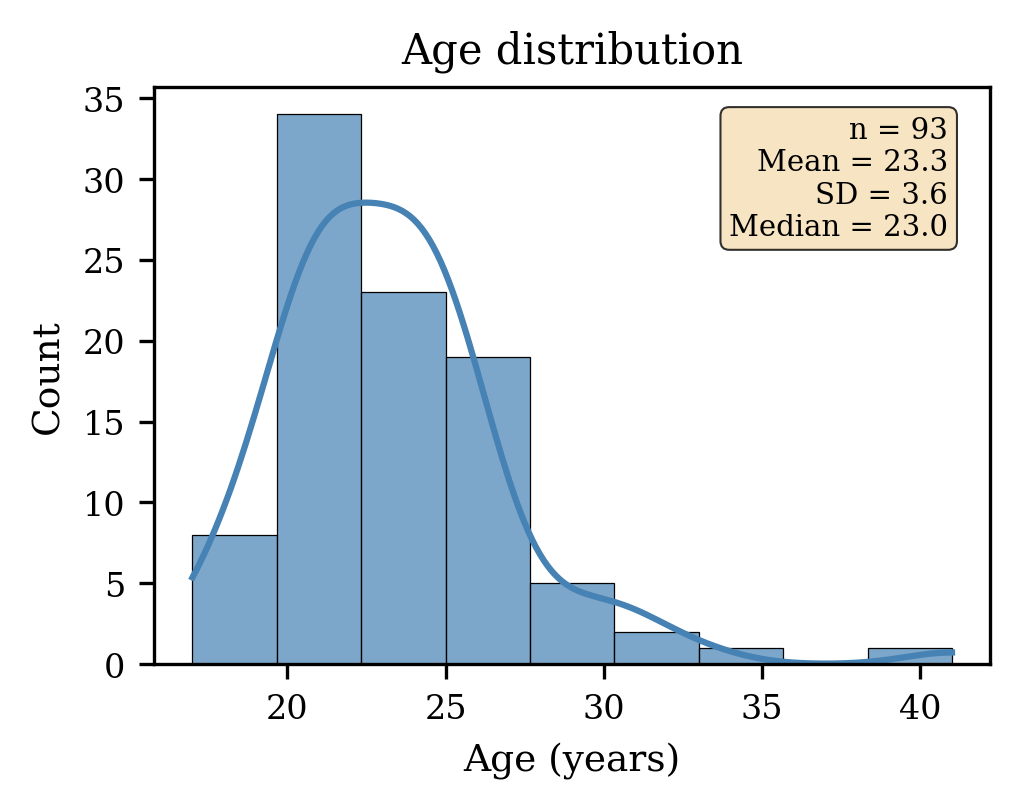}
\caption{Age distribution of the post-cleaning cohort.}
\label{fig:age_dist_post}
\end{figure}

After data cleaning and quality control, the final dataset used for machine learning comprised 93 urine samples, of which 80\% were PCR-positive for \textit{Chlamydia trachomatis}. The post-cleaning cohort included 68 female participants (73.1\%) and 25 male participants (26.9\%). The age distribution of the post-cleaning cohort is shown in Fig~\ref{fig:age_dist_post}.

\subsection{Machine Learning Models}

Five supervised learning models were evaluated to represent a range of linear, nonlinear, ensemble-based, and distance-based decision mechanisms while maintaining conservative model complexity appropriate for a small clinical dataset. All models were implemented using \texttt{scikit-learn}-style pipelines to ensure consistent preprocessing and to prevent data leakage during cross-validation.

Prior to classification, continuous features were standardized using z-score normalization (zero mean, unit variance) implemented within a preprocessing pipeline. Scaling parameters were learned exclusively from the training data within each cross-validation fold and subsequently applied to the corresponding test data. All other preprocessing steps, including feature encoding and missing data handling, were performed during the data curation stage prior to model training.

The following models were considered:

\begin{itemize}
    \item \textbf{Logistic Regression (LR):} A regularized logistic regression model with an L2 penalty was used as a transparent linear baseline. The regularization strength was fixed to the default value ($C=1.0$), and class imbalance was addressed using balanced class weights. The \texttt{liblinear} solver was used with a fixed random seed.

    \item \textbf{Decision Tree (DT):} A shallow decision tree classifier was included to model simple nonlinear interactions. Tree depth was limited to a maximum of four levels, with a minimum of five samples per leaf node. Balanced class weights were applied to mitigate class imbalance.

    \item \textbf{Random Forest (RF):} An ensemble of 200 decision trees was trained to capture nonlinear relationships while improving robustness relative to a single tree. Individual tree depth and minimum leaf size were constrained to match the decision tree baseline. Balanced class weights were applied.

    \item \textbf{Extreme Gradient Boosting (XGB):} A gradient-boosted tree model was evaluated to assess whether modest boosting could improve discrimination beyond bagged ensembles. The number of boosting rounds was fixed to 200, with limited tree depth (maximum depth of three), a learning rate of 0.1, and subsampling of both samples and features to reduce overfitting. All hyperparameters were fixed \textit{a priori}.

    \item \textbf{k-Nearest Neighbors (KNN):} A distance-based classifier was included as a non-parametric comparator. The number of neighbors was fixed to seven, and predictions were weighted by inverse distance.
\end{itemize}

Hyperparameters for all models were selected conservatively based on common practice and were not optimized using the dataset. This design choice reflects the exploratory nature of the study and prioritizes robustness, interpretability, and reproducibility over maximal predictive performance.

\subsection{Evaluation Protocol}

Model performance was evaluated using stratified 5-fold cross-validation with shuffled folds and a fixed random seed of 42 to ensure reproducibility. For each combination of feature set (F1--F3) and classifier, out-of-fold (OOF) predictions were generated using cross-validated probability estimates. Predicted class probabilities for the positive PCR-confirmed outcome were obtained for each sample from models trained exclusively on the remaining folds. Binary predictions were derived using a fixed probability threshold of 0.5.

To prevent data leakage, all preprocessing steps performed within the modeling pipelines, including feature standardization, were fitted only on the training data within each cross-validation split and subsequently applied to the corresponding test fold. No model selection or hyperparameter optimization was performed within the cross-validation procedure.

\subsection{Evaluation Metrics}

Let $y_i \in \{0,1\}$ denote the PCR-confirmed infection status (ground truth) and $\hat{y}_i \in \{0,1\}$ the predicted label obtained by thresholding the predicted probability $\hat{p}_i$ at a fixed threshold of 0.5. Define true positives (TP), true negatives (TN), false positives (FP), and false negatives (FN) as:
\begin{equation}
\begin{aligned}
\mathrm{TP} &= \sum_i \mathbb{1}(y_i = 1 \land \hat{y}_i = 1), \\
\mathrm{TN} &= \sum_i \mathbb{1}(y_i = 0 \land \hat{y}_i = 0), \\
\mathrm{FP} &= \sum_i \mathbb{1}(y_i = 0 \land \hat{y}_i = 1), \\
\mathrm{FN} &= \sum_i \mathbb{1}(y_i = 1 \land \hat{y}_i = 0).
\end{aligned}
\end{equation}

Sensitivity, specificity, precision, and F1-score are defined as:
\begin{equation}
\mathrm{Sensitivity} = \frac{\mathrm{TP}}{\mathrm{TP} + \mathrm{FN}},
\end{equation}
\begin{equation}
\mathrm{Specificity} = \frac{\mathrm{TN}}{\mathrm{TN} + \mathrm{FP}},
\end{equation}
\begin{equation}
\mathrm{Precision} = \frac{\mathrm{TP}}{\mathrm{TP} + \mathrm{FP}},
\end{equation}
\begin{equation}
\mathrm{F1} = 2 \cdot \frac{\mathrm{Precision} \cdot \mathrm{Sensitivity}}{\mathrm{Precision} + \mathrm{Sensitivity}}.
\end{equation}

The area under the receiver operating characteristic curve (AUC) is defined as:
\begin{equation}
\mathrm{AUC} = \Pr\left( \hat{p}^{+} > \hat{p}^{-} \right),
\end{equation}
where $\hat{p}^{+}$ and $\hat{p}^{-}$ denote predicted probabilities for randomly selected positive and negative samples, respectively.

Bootstrap confidence intervals were estimated by resampling the out-of-fold predictions with replacement. For a metric $m$, the $100(1-\alpha)\%$ confidence interval is given by:
\begin{equation}
\left[ Q_{\alpha/2}(m^*),\; Q_{1-\alpha/2}(m^*) \right],
\end{equation}
where $Q_q(\cdot)$ denotes the empirical $q$-quantile of the bootstrap distribution. Bootstrap resampling was performed on the aggregated out-of-fold predictions to avoid optimistic bias.

\section{Results}
\subsection{Predictive Performance Across Feature Sets}

Models trained on patient-reported history and symptoms (F1) demonstrated moderate but variable discriminative performance across classifiers (Table~\ref{tab:results_f1}). While individual models achieved peak AUC values up to approximately 0.72, performance differed substantially between classifiers and confidence intervals were wide. Precision and F1-scores were generally modest, indicating limited robustness and sensitivity of patient-reported features when used in isolation.

Models trained on urine biomarkers (F2) exhibited slightly lower peak discrimination but more consistent performance across classifiers (Table~\ref{tab:results_f2}). AUC values were generally moderate, with ensemble-based methods yielding the most stable results. Compared to F1, biomarker-based models showed reduced variability across classifiers and more balanced threshold-dependent metrics, suggesting that urine biomarkers encode a reliable biological signal associated with infection status.

The combined feature set (F3) yielded stable and competitive performance across models (Table~\ref{tab:results_f3}). While integration of patient-reported features with urine biomarkers did not result in a substantial or consistent increase in peak AUC relative to the best-performing single feature sets, performance variability across classifiers was reduced. Ensemble methods again achieved the strongest overall discrimination, indicating that feature integration primarily enhances robustness rather than maximal discriminative ability.

\begin{table}[htbp]
\caption{Performance for F1 (patient-reported history and symptoms). Metrics are computed from out-of-fold predictions; 95\% confidence intervals are bootstrap-based (1{,}000 resamples).}
\label{tab:results_f1}
\centering
\resizebox{\columnwidth}{!}{%
\begin{tabular}{|c|c|c|c|}
\hline
\textbf{Model} & \textbf{AUC [95\% CI]} & \textbf{Precision [95\% CI]} & \textbf{F1 [95\% CI]} \\
\hline
LR  & 0.6726 [0.5557, 0.7923] & 0.395 [0.229, 0.556] & 0.417 [0.258, 0.556] \\
DT  & 0.6252 [0.4852, 0.7737] & 0.375 [0.227, 0.526] & 0.405 [0.257, 0.537] \\
RF  & 0.7207 [0.5835, 0.8725] & 0.343 [0.194, 0.500] & 0.348 [0.203, 0.479] \\
XGB & 0.5596 [0.4088, 0.7161] & 0.310 [0.154, 0.474] & 0.286 [0.147, 0.432] \\
KNN & 0.6007 [0.4389, 0.7688] & 0.619 [0.391, 0.842] & 0.473 [0.292, 0.630] \\
\hline
\end{tabular}%
}
\end{table}

\begin{table}[htbp]
\caption{Performance for F2 (urine biomarkers). Metrics are computed from out-of-fold predictions; 95\% confidence intervals are bootstrap-based (1{,}000 resamples).}
\label{tab:results_f2}
\centering
\resizebox{\columnwidth}{!}{%
\begin{tabular}{|c|c|c|c|}
\hline
\textbf{Model} & \textbf{AUC [95\% CI]} & \textbf{Precision [95\% CI]} & \textbf{F1 [95\% CI]} \\
\hline
LR  & 0.4881 [0.3478, 0.6360] & 0.676 [0.524, 0.814] & 0.704 [0.567, 0.815] \\
DT  & 0.6237 [0.4634, 0.7897] & 0.558 [0.411, 0.686] & 0.674 [0.541, 0.769] \\
RF  & 0.6548 [0.4852, 0.8271] & 0.625 [0.469, 0.763] & 0.676 [0.540, 0.783] \\
XGB & 0.6496 [0.4865, 0.8164] & 0.710 [0.533, 0.852] & 0.677 [0.533, 0.790] \\
KNN & 0.5785 [0.4292, 0.7398] & 0.652 [0.450, 0.840] & 0.526 [0.359, 0.667] \\
\hline
\end{tabular}%
}
\end{table}

\begin{table}[htbp]
\caption{Performance for F3 (combined patient-reported and urine biomarkers). Metrics are computed from out-of-fold predictions; 95\% confidence intervals are bootstrap-based (1{,}000 resamples).}
\label{tab:results_f3}
\centering
\resizebox{\columnwidth}{!}{%
\begin{tabular}{|c|c|c|c|}
\hline
\textbf{Model} & \textbf{AUC [95\% CI]} & \textbf{Precision [95\% CI]} & \textbf{F1 [95\% CI]} \\
\hline
LR  & 0.5993 [0.4511, 0.7515] & 0.571 [0.400, 0.743] & 0.580 [0.424, 0.711] \\
DT  & 0.5967 [0.4626, 0.7433] & 0.528 [0.392, 0.660] & 0.644 [0.518, 0.753] \\
RF  & 0.7244 [0.5835, 0.8725] & 0.567 [0.385, 0.741] & 0.531 [0.377, 0.667] \\
XGB & 0.6711 [0.5189, 0.8293] & 0.714 [0.531, 0.871] & 0.645 [0.491, 0.769] \\
KNN & 0.6319 [0.5011, 0.7702] & 0.375 [0.000, 0.750] & 0.143 [0.000, 0.286] \\
\hline
\end{tabular}%
}
\end{table}

\subsection{Model Comparison}

\begin{figure}[htbp]
\centering
\includegraphics[width=\columnwidth]{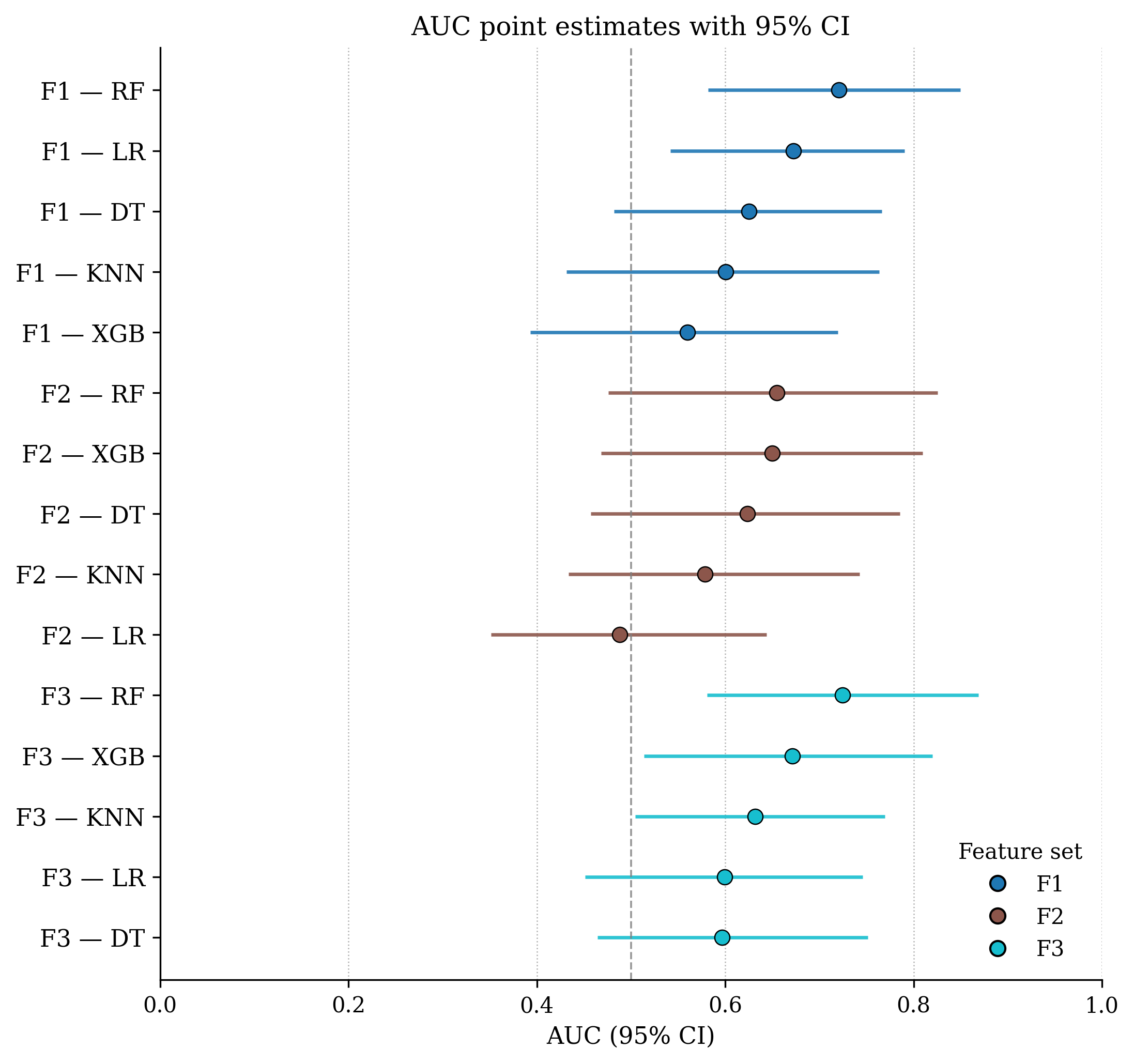}
\caption{AUC point estimates with 95\% bootstrap confidence intervals for all models and feature sets. Vertical dashed line indicates AUC = 0.5 (no-discrimination). Results are computed from out-of-fold predictions to avoid optimistic bias.}
\label{fig:auc_ci}
\end{figure}

Across feature sets, ensemble-based classifiers generally achieved stronger and more consistent performance than linear and distance-based models. As shown in Fig.~\ref{fig:auc_ci}, random forest and gradient-boosted tree models yielded the highest and most stable AUC point estimates across feature sets, particularly for models incorporating urine biomarkers (F2 and F3). While models trained on patient-reported features alone (F1) achieved comparable peak AUC values in individual ensemble models, their performance exhibited greater variability and wider confidence intervals across classifiers.

Logistic regression demonstrated competitive performance for patient-reported feature sets despite its simpler structure, achieving moderate discrimination with relatively stable estimates. In contrast, its performance was weaker for urine biomarker–based models, suggesting limited capacity to capture nonlinear relationships present in biomarker data. Single decision trees showed moderate discrimination across feature sets but were consistently outperformed by ensemble methods, reflecting the benefit of aggregation for capturing nonlinear effects and improving robustness.

\begin{figure}[htbp]
\centering
\includegraphics[width=\columnwidth]{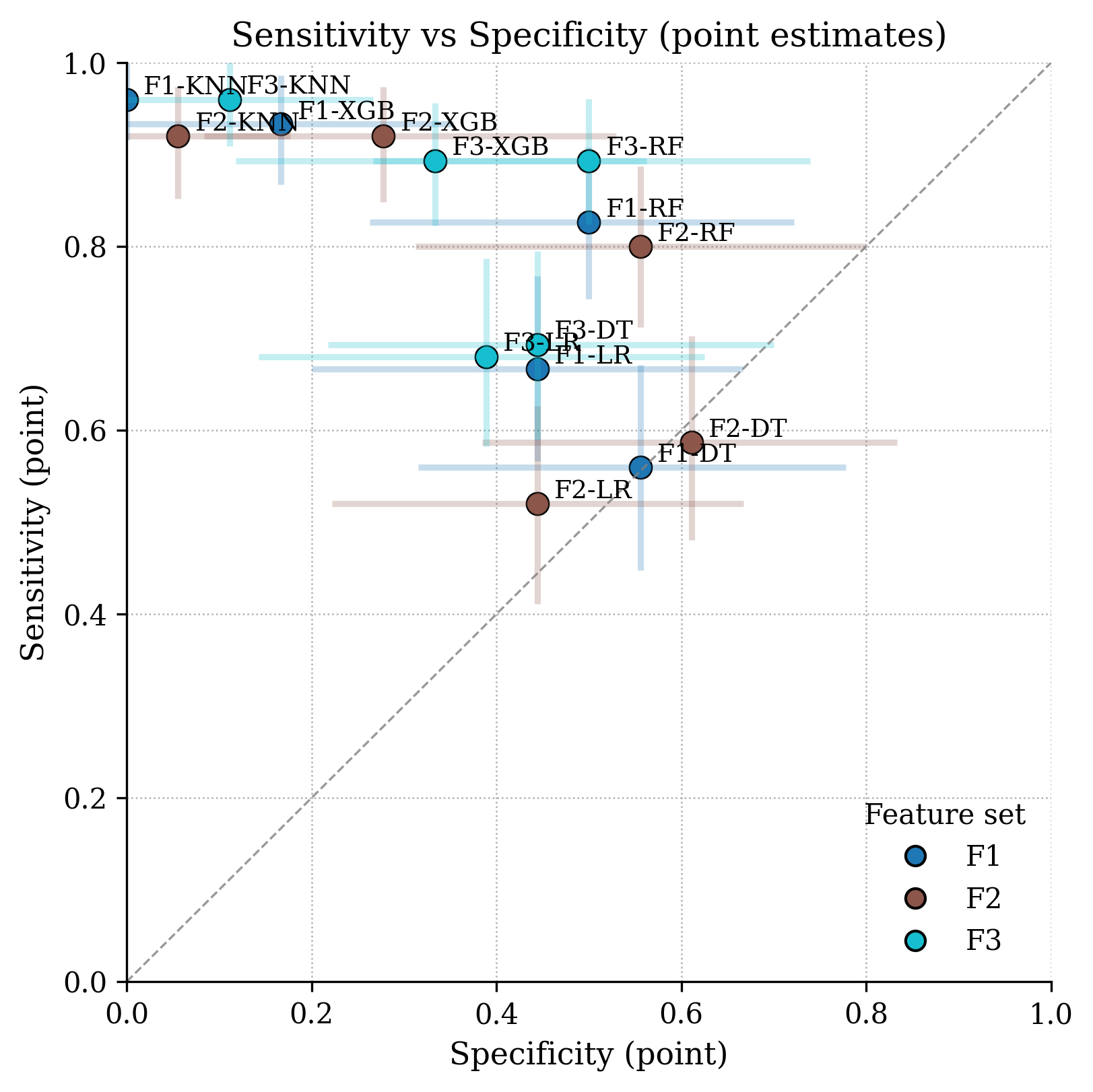}
\caption{Sensitivity versus specificity for all models and feature sets based on point estimates from out-of-fold predictions. Each marker represents a model–feature set combination (F1–F3), with error bars indicating bootstrap uncertainty. The diagonal line denotes equal sensitivity and specificity.}
\label{fig:sens_spec}
\end{figure}

The sensitivity–specificity trade-offs illustrated in Fig.~\ref{fig:sens_spec} further highlight differences between model families. Models incorporating urine biomarkers generally occupied more stable operating regions, achieving balanced sensitivity and specificity across classifiers. Ensemble models tended to exhibit more favorable and consistent trade-offs, whereas distance-based models showed greater variability and uncertainty, particularly for the combined feature set (F3).

\section{Discussion}
This study investigated the feasibility of predicting PCR-confirmed Chlamydia trachomatis infection using machine learning models trained on patient-reported history and symptoms, urine biomarkers, and their combination. The results reveal meaningful differences in predictive behavior across feature sets and provide insight into how distinct sources of non-invasive clinical information contribute to model performance in a pre-test risk stratification (PTRS) context.

Models trained exclusively on patient-reported features (F1) demonstrated moderate but variable discriminative ability across classifiers. While individual models—particularly ensemble methods—achieved peak AUC values up to approximately 0.72, performance varied substantially between classifiers and confidence intervals were wide. Precision and F1-scores were generally modest, indicating limited robustness when patient-reported information was used in isolation. These findings are consistent with the frequently asymptomatic or nonspecific clinical presentation of chlamydia and suggest that patient-reported data alone may be insufficient for reliable and stable prediction, while still capturing contextual risk factors relevant to infection.

In contrast, models trained on urine biomarkers (F2) exhibited more consistent discriminative performance across classifiers, despite achieving slightly lower peak AUC values compared to the best-performing patient-reported models. Ensemble-based approaches, including random forest and gradient-boosted trees, yielded stable AUC estimates and more balanced threshold-dependent metrics. This consistency suggests that routinely collected urine biomarkers encode a reliable biological signal associated with infection status. Importantly, these patterns emerged using conservative, fixed hyperparameters, supporting the robustness of the observed trends despite the limited sample size and reinforcing the exploratory nature of the analysis.

The combined feature set (F3), integrating patient-reported information with urine biomarkers, yielded stable and competitive performance across model families. Although feature integration did not result in a substantial or consistent increase in peak AUC relative to the best-performing single feature sets, performance variability across classifiers was reduced. This pattern indicates that patient-reported information may provide complementary contextual value that stabilizes model predictions when used alongside objective biomarker data, even if it is insufficient as a standalone predictive signal.

From a modeling perspective, ensemble-based classifiers generally achieved stronger and more consistent discrimination than linear and distance-based methods, particularly for feature sets incorporating biomarkers. This aligns with the expectation that nonlinear interactions among clinical variables may be relevant for infection prediction. Nevertheless, logistic regression remained competitive for patient-reported feature sets, highlighting that simpler and more interpretable models may still be viable in PTRS settings when informative features are available and careful evaluation protocols are applied.

Several limitations should be considered when interpreting these findings. The dataset size is modest, limiting statistical power and generalizability, and all results should therefore be interpreted as exploratory and hypothesis-generating. No external validation was performed, and analyses were restricted to a single aggregated dataset. In addition, binary decision thresholds were fixed and not optimized for specific clinical operating points, which may influence reported sensitivity and specificity.

Some models exhibited relatively favorable threshold-dependent metrics despite moderate AUC values, reflecting the fixed probability threshold and underlying class prevalence. While AUC captures global ranking ability, precision and F1-score depend on the selected operating point and should be interpreted accordingly, particularly in triage-oriented screening contexts.

Despite these limitations, this study emphasizes the importance of rigorous evaluation design in small-sample clinical machine learning, including the use of out-of-fold predictions, conservative model complexity, and bootstrap-based uncertainty estimation. By prioritizing transparency, reproducibility, and interpretability, this work provides a clear and reproducible baseline for future investigations using larger cohorts and prospective validation.

Overall, the findings suggest that urine biomarkers provide a reliable and robust standalone signal for PTRS, while patient-reported features can achieve comparable peak discrimination in individual models but with greater variability. Integrating patient-reported information with biomarker data primarily enhances robustness rather than maximal discrimination. These results motivate further investigation into integrated, non-invasive clinical feature representations as upstream decision-support components to complement molecular testing in resource-aware and decentralized screening workflows, including home-based PCR contexts.

\section{Conclusion}

This study explored the feasibility of pre-test risk stratification for PCR-confirmed Chlamydia trachomatis infection using conservative machine learning models trained on patient-reported information, urine biomarkers, and their combination. Across evaluated feature sets and classifiers, distinct differences in predictive behavior were observed, highlighting how different sources of non-invasive clinical data contribute to risk stratification performance.

Patient-reported history and symptoms alone demonstrated moderate but variable discriminative ability, reflecting the frequently asymptomatic and nonspecific clinical presentation of chlamydia. While individual models achieved relatively high peak discrimination, performance was inconsistent across classifiers. Urine biomarkers, in contrast, provided a more reliable and consistent predictive signal across model families, achieving moderate discrimination without extensive hyperparameter tuning. Integrating patient-reported information with biomarkers did not consistently increase peak performance but yielded more stable and competitive results across classifiers, suggesting complementary contextual value that enhances robustness.

Importantly, this work does not aim to replace molecular diagnostics, but rather to assess the relative contribution of different pre-test information sources under a controlled and leakage-aware evaluation framework. By employing predefined feature sets, conservative model complexity, out-of-fold prediction, and bootstrap-based uncertainty estimation, the analysis prioritizes transparency, robustness, and reproducibility under small-sample constraints.

While the findings are exploratory and limited by dataset size and the absence of external validation, they provide a reproducible baseline for future investigations. Further work using larger and more diverse cohorts, prospective validation, and clinically motivated operating points is required to assess real-world utility. Overall, the results support the potential value of non-invasive, routinely available data—particularly urine biomarkers, alone or in combination with patient-reported information—for informing pre-test risk stratification in resource-aware and decentralized screening contexts. In such settings, these models may serve as upstream decision-support tools to prioritize PCR testing for individuals at elevated predicted risk, rather than to provide definitive diagnosis.

\section*{Acknowledgment}

The authors thank Selfdiagnostics Deutschland GmbH (https://selfdiagnostics.eu/) for providing access to the anonymized dataset used in this study. Data collection and assay development were conducted independently by the company prior to this analysis.

\section*{Data Availability and Ethics Statement}

All data used in this study were anonymized prior to access, and no personally identifiable information was available to the authors. The dataset was generated by Selfdiagnostics Deutschland GmbH and is not publicly available due to data ownership and privacy considerations. Access to the data may be provided for academic research purposes upon reasonable request, subject to approval by the data owner.

\section*{Conflict of Interests}
Marko Lehes and Tamas Pardy are affiliated with Selfdiagnostics Deutschland GmbH, which owns the dataset used in this study. No monetary support was provided; the dataset was supplied free of charge. The authors declare no other conflicts of interest.


\begin{thebibliography}{00}

\bibitem{WHOChlamydiaFactSheet}
World Health Organization.
Chlamydia.
\emph{WHO Fact Sheets}
(World Health Organization, Geneva, 2023).
\url{https://www.who.int/news-room/fact-sheets/detail/chlamydia}


\bibitem{Chang2023ChlamydiaEditorial}
Li, C., Ong, J., Tang, W. \& Wang, C.
Editorial: Chlamydia trachomatis infection: Epidemiology, prevention, clinical, and basic science research.
\emph{Frontiers in Public Health} \textbf{11}, Article 1167690 (Frontiers Media S.A., 2023).
\url{https://www.frontiersin.org/journals/public-health/articles/10.3389/fpubh.2023.1167690}
doi:10.3389/fpubh.2023.1167690


\bibitem{Peeling2017AsymptomaticSTI}
Peeling, R. W., Mabey, D., Kamb, M. L., Chen, X. S., Radolf, J. D. \& Benzaken, A. S.
Sexually transmitted infections: challenges ahead.
\emph{The Lancet Infectious Diseases} \textbf{17}, e235--e279
(Elsevier, 2017).
\url{https://www.sciencedirect.com/science/article/pii/S1473309917300109}

\bibitem{CDC2021ChlamydiaGuidelines}
Centers for Disease Control and Prevention.
Chlamydial Infections.
In \emph{Sexually Transmitted Infections Treatment Guidelines, 2021}
(Centers for Disease Control and Prevention, Atlanta, 2021).
\url{https://www.cdc.gov/std/treatment-guidelines/chlamydia.htm}

\bibitem{Gaydos2017NAATReview}
Gaydos, C. A. \& Van Der Pol, B.
Advances in the laboratory diagnosis of Chlamydia trachomatis infections.
\emph{Clinical Microbiology Reviews} \textbf{30}, 686--709
(American Society for Microbiology, 2017).
\url{https://journals.asm.org/doi/10.1128/CMR.00024-16}

\bibitem{Chakraborty2024DemocratizingNAAT}
Chakraborty, S.
Democratizing nucleic acid-based molecular diagnostic tests for infectious diseases at resource-limited settings—from point of care to extreme point of care.
\emph{Sensors \& Diagnostics} \textbf{3}, 536--561 (Royal Society of Chemistry, 2024).
doi:10.1039/D3SD00304C

\bibitem{WHO2025STIGuidance}
World Health Organization.
WHO expands guidance on sexually transmitted infections and reviews country progress on policy implementation.
World Health Organization News Release, July 26, 2025. Available: https://www.who.int/news/item/26-07-2025-who-expands-guidance-on-sexually-transmitted-infections-and-reviews-country-progress-on-policy-implementation

\bibitem{Bilal2021STIPredictionML}
Bilal, M. et al.
Machine learning prediction of incident sexually transmitted infections among men who have sex with men: a cohort study.
\emph{The Lancet Digital Health} \textbf{3}, e642--e651 (2021).

\bibitem{Hao2025STIRiskPrioritization}
Hao, S., Velásquez, E. E., Pearson, W. S., Hoover, K. W., Zhu, W., Rochlin, I. \emph{et al.}
Primary care screening for sexually transmitted infections in the United States from 2019 to 2021.
\emph{PLoS One} \textbf{20}, e0325097 (2025).
\url{https://doi.org/10.1371/journal.pone.0325097}

\bibitem{vanKlaveren2016ChlamydiaPrediction}
van Klaveren, D. \emph{et al.}
Prediction of Chlamydia trachomatis infection to facilitate selective screening: model development and validation using registry and questionnaire data.
\emph{Sexually Transmitted Infections} \textbf{92}, 433–440 (BMJ Publishing Group, 2016).
\url{https://sti.bmj.com/content/92/6/433}

\bibitem{Jensen2025ChlamydiaRiskFactors}
Jensen, S. C., Hoffmann, S., Nardone, A., Parajuli, A. \emph{et al.}
Technologies, strategies and approaches for testing populations at risk of sexually transmitted infections: a systematic review protocol.
\emph{Systematic Reviews} \textbf{14}, Article 107 (2025).
\url{https://springermedicine.com/chlamydia-trachomatis/sexually-transmitted-infection/technologies-strategies-and-approaches-for-testing-populations-a/23011400}

\bibitem{Sequeira-Antunes2023UrinaryBiomarkers}
Sequeira-Antunes, B. \& Ferreira, H. A.
Urinary biomarkers and point-of-care urinalysis devices for early diagnosis and management of disease: a review.
\emph{Biomedicines} \textbf{11}, 1051 (MDPI, 2023).
\url{https://doi.org/10.3390/biomedicines11041051}

\bibitem{Rajkomar2019MLHealthcare}
Rajkomar, A., Dean, J. \& Kohane, I.
Machine learning in medicine.
\emph{New England Journal of Medicine} \textbf{380}, 1347–1358 (2019).
\url{https://doi.org/10.1056/NEJMra1814259}

\bibitem{Bilal2021MLSTDIPredict}
Bilal, M., Ko, Y. A., King, C., McCoy, J., Doros, G., Zhang, X. \emph{et al.}
Machine learning prediction of incident sexually transmitted infections among men who have sex with men: a cohort study.
\emph{Lancet Digital Health} \textbf{3}, e642–e651 (2021).
\url{https://doi.org/10.1016/S2589-7500(21)00102-6}

\end{thebibliography}
\end{document}